\documentclass[twoside]{article}
\usepackage[normalem]{ulem}
\usepackage{graphicx}
\usepackage[utf8]{inputenc} 
\usepackage[T1]{fontenc}    
\usepackage{hyperref}       
\usepackage{url}            
\usepackage{booktabs}       
\usepackage{amsmath,amssymb,amsfonts,amsthm,bbm}
\usepackage{mathtools}   
\usepackage{stmaryrd} 
\usepackage{enumitem}   
\usepackage{nicefrac}       
\usepackage{microtype}      
\usepackage{xcolor}         
\usepackage{mathrsfs}
\usepackage{comment}
\usepackage{stmaryrd} 

\renewcommand{\hat}{\widehat}
\renewcommand{\tilde}{\widetilde}

\newcommand{\bfm}[1]{\ensuremath{\boldsymbol{#1}}} 

   \def\bA{\bfm A}  
\def\bb{\bfm b}     
     
\def\bd{\bfm d}   \def\bD{\bfm D}  
     \def\EE{\mathbb{E}}

     \def\NN{\mathbb{N}}

     \def\RR{\mathbb{R}}

   \def\bU{\bfm U}

   \def\bX{\bfm X}

\def\calB{{\cal  B}} 
\def\calC{{\cal  C}} 
 
\def\calE{{\cal  E}} 
 
\def\calG{{\cal  G}} 
\def\calH{{\cal  H}}

\def\calL{{\cal  L}}

\def\calT{{\cal  T}}

\def\calW{{\cal  W}} 
\def\calX{{\cal  X}} 
 
\def\calZ{{\cal  Z}}

\def\bcalB{{\boldsymbol{\calB}}}
\def\bcalC{{\boldsymbol{\calC}}}
\def\bcalE{{\boldsymbol{\calE}}}
\def\bcalH{{\boldsymbol{\calH}}}

\def\bcalW{{\boldsymbol{\calW}}}
\def\bcalX{{\boldsymbol{\calX}}}
\providecommand{\abs}[1]{\left\lvert#1\right\rvert}
\providecommand{\norm}[1]{\left\lVert#1\right\rVert}

\providecommand{\angles}[1]{\left\langle #1 \right\rangle}
\providecommand{\paran}[1]{\left( #1 \right)}
\providecommand{\brackets}[1]{\left[ #1 \right]}

\providecommand{\braces}[1]{\left\{ #1 \right\}}

\providecommand{\defeq}{:=}

\usepackage{mathtools}
\DeclarePairedDelimiterX{\infdivx}[2]{(}{)}{%
  #1 \; \delimsize\| \; #2%
}

\DeclareMathOperator{\sgn}{sgn}

\newtheorem{definition}{Definition}[section]

\newtheorem{lemma}[definition]{Lemma}

\newtheorem{theorem}[definition]{Theorem}

\definecolor{royalpurple}{rgb}{0.47, 0.32, 0.66}
\definecolor{greenfresh}{HTML}{00897B}
\definecolor{bluefresh}{HTML}{1E88E5}
\definecolor{redfresh}{HTML}{E53935}

\definecolor{royalpurple}{rgb}{0.47, 0.32, 0.66}

\def\beq{\begin{equation}}
\def\eeq{\end{equation}}

\def\bet{\begin{theorem}}
\def\eet{\end{theorem}}

\def\bel{\begin{lemma}}
\def\eel{\end{lemma}}

\def\cond{\;|\;}

\usepackage{setspace}
\usepackage{etoolbox}
\BeforeBeginEnvironment{equation*}{\begin{singlespace}\vspace*{-\baselineskip}}
	\AfterEndEnvironment{equation*}{\end{singlespace}\noindent\ignorespaces}
\BeforeBeginEnvironment{equation}{\begin{singlespace}\vspace*{-\baselineskip}}
	\AfterEndEnvironment{equation}{\end{singlespace}\noindent\ignorespaces}
\BeforeBeginEnvironment{align}{\begin{singlespace}\vspace*{-\baselineskip}}
	\AfterEndEnvironment{align}{\end{singlespace}\noindent\ignorespaces}
\BeforeBeginEnvironment{align*}{\begin{singlespace}\vspace*{-\baselineskip}}
	\AfterEndEnvironment{align*}{\end{singlespace}\noindent\ignorespaces}
\BeforeBeginEnvironment{eqnarray}{\begin{singlespace}\vspace*{-\baselineskip}}
	\AfterEndEnvironment{eqnarray}{\end{singlespace}\noindent\ignorespaces}
\BeforeBeginEnvironment{eqnarray*}{\begin{singlespace}\vspace*{-\baselineskip}}
	\AfterEndEnvironment{eqnarray*}{\end{singlespace}\noindent\ignorespaces}

\usepackage[accepted]{aistats2024}

\begin{document}

%

%

\twocolumn[

\aistatstitle{Tensor-view Topological Graph Neural Network}

\aistatsauthor{ Tao Wen \And Elynn Chen \And  Yuzhou Chen }

\aistatsaddress{ Center for Data Science \\ New York University \\ \href{mailto:tw2672@nyu.edu}{tw2672@nyu.edu}
 \And  Stern School of Business\\New York University \\ \href{mailto:elynn.chen@stern.nyu.edu}{elynn.chen@stern.nyu.edu} \And Computer and Information Sciences \\ Temple University \\ \href{mailto:yuzhou.chen@temple.edu}{yuzhou.chen@temple.edu}} ]

\begin{abstract}
Graph classification is an important learning task for graph-structured data. 
Graph neural networks (GNNs) have recently gained growing attention in graph learning and shown significant improvements on many important graph problems. 
Despite their state-of-the-art performances, existing GNNs only use local information from a
very limited neighborhood around each node, suffering from loss of multi-modal information and overheads of excessive
computation. 
To address these issues, we propose a novel {\em Tensor-view Topological Graph Neural Network}  (TTG-NN), a class of simple yet effective topological deep learning built upon persistent homology, graph convolution, and tensor operations. 
This new method incorporates {\em tensor learning} to simultaneously capture {\it Tensor-view Topological} (TT), as well as
{\it Tensor-view Graph} (TG) structural information on both local and global levels. 
Computationally, to fully exploit graph topology and structure, we propose two flexible TT and TG representation learning modules that disentangle feature tensor aggregation and transformation, and learn to preserve multi-modal structure with less computation.
Theoretically, we derive high probability bounds on both the out-of-sample and in-sample mean squared approximation errors for our proposed {\it Tensor Transformation Layer} (TTL). 
Real data experiments show that the proposed TTG-NN outperforms 20 state-of-the-art methods on various graph benchmarks.
\end{abstract}

\section{Introduction}

Graph data are ubiquitous: many real-world objects can be represented by graphs, such as images, text, molecules, social networks, and power grids. 
Tremendous advances on graph analysis have been achieved in recent years, especially in the field of machine learning (ML) and deep learning (DL)~\cite{defferrard2016convolutional,bronstein2017geometric,zhang2020deep}. In particular, graph neural networks (GNNs) have emerged as effective architectures for various prediction problems, e.g., node classification~\cite{kipf2016semi,velivckovic2017graph,hamilton2017inductive}, link prediction~\cite{zhang2018link,chen2021bscnets}, graph classification~\cite{xu2018powerful,ying2018hierarchical}, and spatio-temporal forecast~\cite{guo2019attention,zhao2019t,bai2020adaptive}. GNNs are neural network architectures specifically designed to handle graph-structured data. The fundamental idea behind GNNs involves treating the underlying graph as a computation graph and leveraging neural network primitives to generate node embeddings. This process involves transforming, propagating, and aggregating node features and graph structural information throughout the graph. However, most GNNs follow a neighborhood aggregation process where
the feature vector of each node is computed by recursively
aggregating and transforming the representation vectors of its neighbors. Consequently, they are unable to capture higher-order relational structures and local topological information concealed in the graph, which are highly relevant for applications that rely on connectivity information~\cite{you2018graph,huang2020skipgnn,sun2022does}. For instance, understanding the behavior and properties of molecules and protein data within the drug discovery and development process requires capturing crucial information, such as higher-order interactions between atoms, the triangular mesh of a protein surface, and ring-ring interactions within a molecule, etc.

To address these challenges, the integration of ML/DL methods with persistent homology (PH) representations of learned objects have been intensively studied~\cite{wasserman2018topological,carlsson2020topological,o2021filtration,edelsbrunner2022computational}. PH is a methodology under the topological data analysis (TDA) framework that captures the topological features (like connected components, holes, voids, etc.) of a shape at various scales and provides a multi-scale description of the shape. In this case, we can say that PH studies the observed object at multiple resolutions or evaluates topological patterns and structures through multiple lenses. By incorporating this multi-scale topological information, topological-based models can capture both the geometric and topological structures, and gain a more comprehensive representation of the data. Previous research has generated a topological signature and computed a parameterized vectorization that can be integrated into kernel functions~\cite{bubenik2015statistical,kusano2016persistence,reininghaus2015stable}. With growing interest in DL, several topological DL (TDL) methods have been proposed~\cite{hofer2017deep,carriere2020perslay}. Specifically, they extract topological features from underlying data (e.g., topological features encoded in persistence diagrams) and integrate them into any type of DL. Recently,~\cite{zhao2020persistence,chen2021topological,yan2021link,horn2021topological} prove that it is important to
learn node representations based on both the topological structure and node attributes for the graph learning problems. However, these above topological-based models cannot (1) fully capture the rich multi-dimensional/multi-filtrations topological features in objects and (2) exploit the low-rank structure from intermediate layers of TDL models. For instance, topological GNN~\cite{zhao2020persistence,chen2021topological} only calculates the topological features of nodes via a single filtration. In~\cite{horn2021topological,chen2022tamp}, Horn et al. and Chen et al. have addressed the filtration issues, but their methods fail to model topological feature tensors while preserving their low-rank structures.

In this paper, we develop a novel framework, namely Tensor-view Topological Graph Neural Network (TTG-NN) to address the above problems for real-world graph data. More specifically, we propose two novel and effective tensor-based graph representation learning schemes, i.e., Tensor-view Topological Convolutional Layers (TT-CL) and Tensor-view Graph Convolutional Layers (TG-CL). Technically, we first produce topological and structural feature tensors of graphs as 3D or 4D tensors by using {\it multi-filtrations} and graph convolutions respectively. Then, we utilize TT-CL and TG-CL to learn hidden local and global topological representations of graphs.

A naive aggregation of multiple feature tensors will increase the complexity of NN and incur excessive computational costs. We carefully design a module of {\em Tensor Transformation Layers} (TTL) which employs tensor low-rank decomposition to address the model complexity and computation issues. 
By smartly combining the three modules of TT-CL, TC-GL, and TTL, we can safely incorporate multiple topological and graph features without losing any potential discriminant features, and, at the same time, enjoy a parsimonious NN architecture from the low-rankness of the input feature tensors. 
The advantages of our TTG-NN are validated both theoretically by Theorem \ref{thm:oracle} and empirically through our extensive experiments. 

In short, our main contributions are as follows:
(1) This is the first approach bridging tensor methods with an aggregation of multiple features constructed by persistent homology and graph convolution.
(2) We provide the first non-asymptotic error bounds of both in-sample and out-of-sample mean squared errors of TTL with Tucker-low-rank feature tensors. 
(3) Our extensive experiments of TTG-NN on graph classification tasks show that TTG-NN delivers state-of-the-art classification performance with a notable margin, and demonstrates high computational efficiency.

\subsection{Related Work}
\textbf{Graph Neural Networks.} Recently, Graph Neural Network (GNN) has emerged as a primary tool for graph classification~\cite{zhou2020graph, xia2021graph, 10.1145/3495161, 9039675, 10.1613/jair.1.14768}. Different methods have been proposed to capture the structural and semantic properties of graphs.
For instance, Weisfeiler–Lehman (WL)~\cite{shervashidze2011weisfeiler} proposes an efficient family of kernels for large graphs with discrete node labels, and Shortest Path Hash Graph Kernel (HGK-SP)~\cite{morris2016faster} derives kernels for graphs with continuous attributes from discrete ones. Graph Convolutional Network (GCN)~\cite{kipf2016semi} extends the convolution operation from regular grids to graphs. To handle large-scale graphs, Top-$K$ pooling operations~\cite{cangea2018towards, gao2019graph} design a pooling method by using node features and local structural information to propagate only the top-$K$ nodes with the highest scores at each pooling step. To leverage topological information, Topological Graph Neural Networks (TOGL)~\cite{horn2021topological} proposes a layer that incorporates global topological information of a graph using persistent homology and can be integrated into any type of GNN. A common limitation is that they fail to accurately capture higher-order and local topological properties of graphs or incorporate rich structure information both in local and global domains.

\textbf{Tensor-input Neural Networks.} Neural Networks that take tensors as inputs are designed to process and analyze data in a tensor format, allowing for the efficient processing of high-dimensional data. A tensor analysis on the expressive power of deep neural networks~\cite{cohen2016expressive} derives a deep network architecture based on arithmetic circuits that inherently employs locality, sharing and pooling, and establishes an equivalence between neural networks and hierarchical tensor factorizations. Tensor Contraction Layer(TCL)~\cite{kossaifi2017tcl} incorporates tensor contractions as end-to-end trainable neural network layers, regularizes networks by imposing low-rank constraints on the activations, and demonstrates significant model compression without significant impact on accuracy. Tensor Regression Layer(TRL)~\cite{kossaifi2020trl} further regularizes networks by regression weights, and reduces the number of parameters while maintaining or increasing accuracy. Graph Tensor Network(GTN)~\cite{xu2023graph} introduces a Tensor Network-based framework for describing neural networks through tensor mathematics and graphs for large and multi-dimensional data. In conclusion, current Neural Networks with tensor inputs lack both theoretical and empirical in-depth study.

\section{Preliminaries}\label{sec_preliminaries}
\textbf{Problem Setting} Let $\mathcal{G} = (\mathcal{V}, \mathcal{E}, \bX)$ be an attributed graph, where $\mathcal{V}$ is a set of nodes ($|\mathcal{V}|=N$), $\mathcal{E}$ is a set of edges, and $\bX \in \mathbb{R}^{N \times F}$ is a feature matrix of nodes (here $F$ is the dimension of the node features). 
Let $\bA \in \mathbb{R}^{N \times N}$ be a symmetric adjacency matrix whose entries are
$a_{ij} = \omega_{ij}$ if nodes $i$ and $j$ are connected and 0 otherwise 
(here $\omega_{ij}$ is an edge weight and $\omega_{ij}\equiv 1$ for unweighted graphs).
Furthermore, $\bD$ represents the degree matrix of $\bA$, that is $d_{ii} = \sum_j a_{ij}$. 
In the graph classification setting, we have a set of graphs $\{\mathcal{G}_1, \mathcal{G}_2, \dots, \mathcal{G}_\aleph\}$, where each graph $\mathcal{G}_i$ is associated with a label $y_i$. The goal of the graph classification task is to take the graph as the input and predict its corresponding label.

\textbf{Persistent Homology} Persistent Homology (PH) is a subfield of algebraic topology which provides a way for measuring topological features of shapes and functions. These shape patterns represent topological properties such as 0-dimensional topological features (connected components), 1-dimensional topological features (cycles), 2-dimensional topological features (voids), and, in general, $q$-dimensional ``holes'' represent the characteristics of the graph $\mathcal{G}$ that remain preserved at different resolutions under continuous transformations (where $q = \{0, 1, \dots, \mathcal{Q}\}$ and $\mathcal{Q}$ denotes the maximum dimension of the simplicial complex). Through the use of this multi-resolution scheme, PH tackles the inherent restrictions of traditional homology, enabling the extraction of latent shape characteristics of $\mathcal{G}$ which may play an essential role in a given learning task. The key is to select a suitable scale parameter $\epsilon$ and then to study changes in the shape of $\mathcal{G}$ that occur as $\mathcal{G}$ evolves to $\epsilon$. 
Thus, given an increasing sequence $\epsilon_1 < \cdots < \epsilon_n$, we no longer study $\mathcal{G}$ as a single object but as a {\it filtration} $\mathcal{G}_{\epsilon_1} \subseteq \ldots \subseteq \mathcal{G}_{\epsilon_n}=\mathcal{G}$. To ensure that the process of pattern selection and count are objective and efficient, we build an abstract simplicial complex $\mathscr{C}(\mathcal{G}_{\epsilon_j})$ on each $\mathcal{G}_{\epsilon_j}$, which results in filtration of complexes $\mathscr{C}(\mathcal{G}_{\epsilon_1}) \subseteq \ldots \subseteq \mathscr{C}(\mathcal{G}_{\epsilon_n})$. 
For instance, we consider a function on a node set $\mathcal{V}$. That is, we choose a very simple filtration based on the {\it node degree}, i.e., the number of edges that are incident to a node $u \in \mathcal{V}$, and get a descriptor function (i.e., filtration function) $f(u) = \deg{(u)}$. When scanning $\mathcal{G}$ via the degree-based filtration function $f$, it results in a sequence of induced subgraphs of $\mathcal{G}$ with a maximal degree of $\epsilon_j$ for each $j\in\{1,\ldots, n\}$. 
A standard descriptor of the above topological evolution is {\it Persistence Diagram} (PD)~\cite{article} $Dg = \{(b_\rho, d_\rho) \in \mathbb{R}^2 | b_\rho < d_\rho\}$, which is a multi-set of points in $\mathbb{R}^2$. Each persistence point $(b_\rho, d_\rho)$ corresponds to the lifespan (i.e., $d_\rho - b_\rho$) of one topological feature, where $b_\rho$ and $d_\rho$ represent the birth and death time of the topological feature $\rho$.

\section{Methodology: Tensor-view Topological Graph Neural Network}
\begin{figure}[h]
    \includegraphics[width=.5\textwidth]{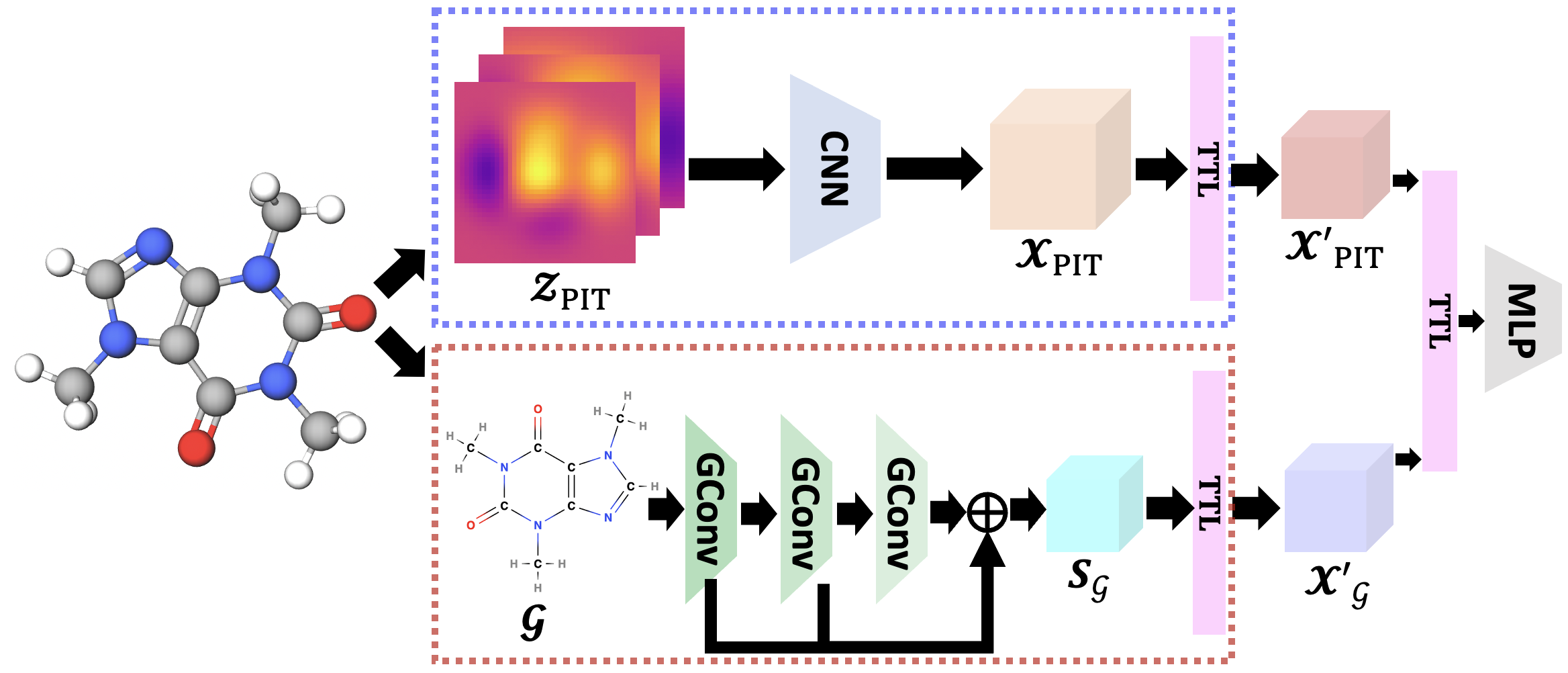}
\vspace{-3ex}
\caption{The architecture of TTG-NN.\label{ttn_nets_flowchart}}
\end{figure}

In this section, we introduce our Tensor-view Topological Graph Neural Network, dubbed as TTG-NN. Our proposed TTG-NN framework is summarized in Figure~\ref{ttn_nets_flowchart}. As illustrated in Figure~\ref{ttn_nets_flowchart}, our method consists of two components. First, tensor-view topological features are extracted by multi-filtrations from multiple views of a graph, and then we design a tensor-view topological representation learning module ({\it Top}) for embedding tensor-view local topological features into a high-dimensional space. Second, we develop a tensor-view graph convolutional module ({\it Bottom}) on a graph to generate a global shape descriptor.

\subsection{Tensor-view Topological Convolutional Layers (TT-CL)}
Our first representation learning module utilizes multiple topological features simultaneously by combining the persistent homology and the proposed tensor learning method. 
To capture the underlying topological features of a graph $\calG$,  
we employ $K$ vertex filtration functions: $f_i: \mathcal{V} \mapsto \mathbb{R}$ for $i\in\{1, \dots, K\}$. 
Each filtration function $f_i$ gradually reveals one specific topological structure at different levels of connectivity, such as the number of relations of a node (i.e., degree centrality score) , node flow information (i.e., betweenness centrality score), information spread capability (i.e., closeness centrality score), and other node centrality measurements. 
With each filtration function $f_i$, we construct a set of $Q$ persistence images of resolution $P \times P$ using tools in persistent homology analysis. 

Combining $Q$ persistence images of resolution $P\times P$ from $K$ different filtration functions, we construct a {\it tensor-view} topological representation, namely {\em Persistent Image (PI) Tensor} $\boldsymbol{\calZ}_\text{PIT}$ of dimension $K \times Q \times P \times P$. 
We design the {\em Tensor-view Topological Convolutional Layer (TT-CL)} to (i) jointly extract and learn the latent topological features contained in the $\boldsymbol{\calZ}_\text{PIT}$, (ii) leverage and preserve the multi-modal structure in the $\boldsymbol{\calZ}_\text{PIT}$, and (iii) capture the structure in trainable weights (with fewer parameters). 
Firstly, hidden representations of the PI tensor $\boldsymbol{\calZ}_\text{PIT}$ are encoded by a combination of a CNN-based neural network and global pooling layers. 
Mathematically, we obtain a learnable topological tensor representation defined by
\begin{equation}\label{gcn_tda_part}
    \boldsymbol{\calX}_{\text{PIT}} =
    \begin{cases}
    f_{\text{CNN}}(\boldsymbol{\calZ}_\text{PIT}) \quad \text{if} \; |Q|=1\\
    \xi_{\text{POOL}}(f_{\text{CNN}}(\boldsymbol{\calZ}_\text{PIT})) \quad \text{if} \; |Q| > 1
    \end{cases},
\end{equation}
where $f_{\text{CNN}}$ is a CNN-based neural network, 
$\xi_{\text{POOL}}$ is a pooling layer that preserves the information of the input in a fixed-size representation (in general, we consider either global average pooling or global max pooling). 
Equation \eqref{gcn_tda_part} provides two simple yet effective methods to extract {\it learnable} topological features: (i) if only considering $q$-dimensional topological features in $\boldsymbol{\calZ}_{\text{PIT}}$, we can apply any CNN-based model to learn the latent feature of the $\boldsymbol{\calZ}_{\text{PIT}}$; 
(ii) if considering topological features with $Q$ dimensions, we can additionally employ a global pooling layer over the latent feature and obtain an image-level feature. 
Secondly, $\boldsymbol{\calX}_{\text{PIT}}$ is fed into our {\em Tensor Transformation Layer (TTL)} as $\boldsymbol{\calH}^{(0)} = \boldsymbol{\calX}_{\text{PIT}}$, whose $\ell$-th layer is defined in \eqref{trl_eq}. 
The output of TTL in TT-CL is denoted as $\boldsymbol{\calX}'_{\text{PIT}}$, which captures the local topological information of a graph. 

\subsection{Tensor-view Graph Convolutional Layers (TG-CL)}

Parallel to the TT-CL is our second representation learning module, {\em Tensor-view Graph Convolutional Layer (TG-CL)}. It utilizes the graph structure of ${\mathcal{G}}$ with its node feature matrix $\bX$ through the graph convolution operation and a multi-layer perceptron (MLP). 
Specifically, the graph convolution operation proceeds by multiplying the input of each layer with the $\tau$-th power of the normalized adjacency matrix. 
The $\tau$-th power operator contains statistics from the $\tau$-th step of a random walk on the graph, thus nodes can indirectly receive more information from farther nodes in the graph. {Unlike~\cite{pmlr-v162-choromanski22a}, different $\tau$-th steps of random walk on the graph are allowed to combine thanks to our tensor architecture, which can enhance the representation power of GCN.}

Combined with an MLP, the representation learned at the $\ell$-th layer is given by
\begin{align}
\label{GNN_MLP_eq}
    {\boldsymbol{S}_{\calG}^{(\ell +1)}} = f_{\text{MLP}}\paran{\varphi\paran{\hat{\bA}^{\tau}\boldsymbol{S}_{\calG}^{(\ell)}\boldsymbol{\Theta}^{(\ell)}}},
\end{align}
where $\hat{\bA} = \boldsymbol{\Tilde{{D}}}^{-\frac{1}{2}}\boldsymbol{\Tilde{{A}}}\boldsymbol{\Tilde{{D}}}^{\frac{1}{2}}$, 
$\boldsymbol{\Tilde{{A}}} = \bA + \boldsymbol{I}$, 
and $\boldsymbol{\Tilde{{D}}}$ is the degree matrix of $\boldsymbol{\Tilde{{A}}}$. $\boldsymbol{S}_{\calG}^{(0)} = \bX$, $f_{\text{MLP}}$ is an MLP with batch normalization, $\varphi (\cdot)$ is a non-linear activation function, $\boldsymbol{\Theta}^{(\ell)}$ is a trainable weight of $\ell$-th layer.
To exploit multi-hop propagation information and increase efficiency, we apply our proposed {\em Tensor Transformation Layer (TTL)} defined in \eqref{trl_eq} over an aggregation of the outputs of all layers in Equation~\eqref{GNN_MLP_eq} to provide structure-aware representations of the input graph. 
Specifically, we first concatenate all layers of a $L$-layer graph convolutions 
$\left[\boldsymbol{S}_{\calG}^{(1)}, \boldsymbol{S}_{\calG}^{(2)}, \dots, \boldsymbol{S}_{\calG}^{(L)}\right]$
to form a node embedding tensor denoted by $\boldsymbol{\calX}_{\mathcal{G}}$ of dimension ${N \times L \times D \times D}$; then $\boldsymbol{\calX}_{\mathcal{G}}$
is fed into TTL as $\boldsymbol{\calH}^{(0)} = \boldsymbol{\calX}_{\mathcal{G}}$, whose $\ell$-th layer is defined in \eqref{trl_eq}. {Note that $\boldsymbol{\calX}_{\mathcal{G}}$ is of dimension ${N \times L \times D \times D}$ so as to facilitate TTL. That is, we prefer both the output tensors of TT-CL and TG-CL to have the same number of dimensions. Suppose the dimension of each node’s representation is $D$ originally, we further conduct $D$ convolutional transformations, which increases the dimension from $D$ to $D^2$, then it is reshaped to $D \times D$.}
The output of TTL in TG-CL is denoted as $\boldsymbol{\calX}'_{\mathcal{G}}$, which captures the global topological information of a graph.  

\subsection{Global and Local Aggregation}

Finally, to aggregate both local and global topological information, we combine representations learned from TT-CL and TG-CL together to obtain the final embedding and feed the concatenated tensor into a TTL then a single-layer MLP for classification. The aggregation operation is defined as
\begin{align*}
    {\boldsymbol{s}}_o = f_\text{MLP}(\text{TTL}([\boldsymbol{\calX}'_\text{PIT}, \boldsymbol{\calX}'_\mathcal{G}])),
\end{align*}
where $\boldsymbol{s}_o$ is the final score matrix for graph classification.

\subsection{Tensor Transformation Layer (TTL)}
The {\em Tensor Transformation Layer (TTL)} preserves the tensor structures of feature $\boldsymbol{\boldsymbol{\calX}}$ of dimension $D=\prod_{m=1}^M D_m$ and hidden throughput.
Let $L$ be any positive integer 
and $\bd=\left[d^{(1)}, \cdots, d^{(L+1)}\right]$ collects the width of all layers. 
A {\em deep ReLU Tensor Neural Network} is a function mapping taking the form of
\begin{equation} \label{eqn:TNN}
    f(\bcalX) = \calL^{(L+1)}\circ\sigma\circ\calL^{(L)}\circ\sigma\cdots\circ\calL^{(2)}\circ\sigma\circ\calL^{(1)}(\bcalX), 
\end{equation}
where $\sigma(\cdot)$ is an element-wise activation function. 
Affine transformation $\calL^{(\ell)}\paran{\cdot}$ and hidden input and output tensor of the $\ell$-th layer, i.e. $\bcalH^{(\ell+1)}$ and $\bcalH^{(\ell)}$ are defined by
\begin{equation} \label{trl_eq}
\begin{aligned}
    \calL^{(\ell)}\paran{\bcalH^{(\ell)}} \defeq \angles{\boldsymbol{\calW}^{(\ell)}, \bcalH^{(\ell)}} + \bcalB^{(\ell)}, \\
    \text{and}\quad
    \bcalH^{(\ell+1)} \defeq \sigma\paran{\calL^{(\ell)}\paran{\bcalH^{(\ell)}}}
\end{aligned}
\end{equation}
where
$\bcalH^{(0)} = \boldsymbol{\boldsymbol{\calX}}$ takes the tensor feature, 
$\angles{\cdot, \cdot}$ is the tensor inner product, 
and {\em low-rank weight} tensor $\boldsymbol{\calW}^{(\ell)}$ and a bias tensor $\bcalB^{(\ell)}$. 
The tensor structure kicks in when we incorporate tensor low-rank structures such as {\it CP low-rank}, {\it Tucker low-rank}, and {\it Tensor Train low-rank}.

Tucker low-rank structure is defined by
\begin{equation} \label{eqn:tucker}
    \bcalX = \bcalC\times_1 \bU_1\times_2\cdots\times_M \bU_M + \bcalE,
\end{equation}
where 
$\bcalE\in\RR^{D_1\times\cdots\times D_M}$ is the tensor of the idiosyncratic component (or noise) and 
$\boldsymbol{\calC}$ is the latent core tensor representing the true low-rank feature tensors and $\bU_m$, $m\in[M]$, are the loading matrices.  

CP low-rank is a special case where the core tensor $\bcalC$ has the same dimensions over all modes, that is $R_m = R$ for all $m\in[M]$, and is super-diagonal. 
TT low-rank is a different kind of low-rank structure, which inherits advantages from both CP and Tucker decomposition.
Specifically, TT decomposition can compress tensors as significantly as CP decomposition, while its calculation is as stable as Tucker decomposition. 


Theoretically, the provable gain of preserving tensor structures and incorporating low-rankness is established in the next section.
Empirically, we perform an ablation study for the effect of different tensor decomposition methods on molecular and chemical graphs. 
Results in Table~\ref{ablation_analysis} align with our theoretical discovery in Theorem \ref{thm:oracle}.

\subsection{Provable Benefits of TTL with Tucker Low-Rankness}

In this section, we show provable benefits from {\em Tensor Transformation Layer} (TTL) with Tucker low-rankness. 
Note that under Tucker low-rankness, it is equivalent to consider either low-rank feature $\boldsymbol{\calX}$ 
or low-rank weight $\boldsymbol{\calW}$. 
Without loss of generality, we consider the low-rankness of feature tensor $\boldsymbol{\calX}$ for theoretical development. 
The feature tensor $\bcalX$ in this section corresponds to the aggregated feature $[\boldsymbol{\calX}'_\text{PIT}, \boldsymbol{\calX}'_\mathcal{G}]$ constructed from TT-CL and TG-CL. 
The theoretical result applies generally to any $M$-th order tensor feature $\bcalX$ of dimension $D_1\times\cdots\times D_M$.

The feature tensors and their corresponding labels $\braces{\boldsymbol{\calX}_i, y_i}_{i=1}^n$ are observable and their corresponding latent cores $\braces{\boldsymbol{\calC}_i}_{i=1}^n$ are i.i.d.~copies of latent $\boldsymbol{\calC}$. 
The underlying true regression model is given by 
\begin{equation}
    \EE\brackets{y \cond \boldsymbol{\calX}} 
    = \EE\brackets{y \cond \boldsymbol{\calC}}
    = m^*(\boldsymbol{\calC}). 
\end{equation}
TTL uses deep ReLU Neural Networks to approximate $m^*(\boldsymbol{\calC})$. 
When $\{\boldsymbol{\calC}_i\}_{i=1}^n$ is not directly observable, TTL estimates it from $\{\boldsymbol{\calX}_i\}_{i=1}^n$ using Tucker decomposition \cite{kolda2009tensor}. 
In this section, we provide the first theoretical guarantee for TTL. 
\begin{definition}[Deep ReLU Tensor Network Class]
For any depth $L\in\NN$, width vector $\bd\in\NN^{L+1}$, $B, M \in \RR^+ \cup \{\infty\}$, the family of deep ReLU Tensor Network truncated by $M$ with depth $L$, width parameter $\bd$, and weights bounded by $B$ is defined as
\begin{equation*}
     \calC(L, \bd, M, B) = \left\{\bar{f}(\boldsymbol{\calX}) = \calT_M(f(\boldsymbol{\calX}))\} \right.
\end{equation*}
where
\begin{equation*}
\begin{aligned}
f(\boldsymbol{\calX}) \text{ defined in \eqref{eqn:TNN} and \eqref{trl_eq} with }\\ \norm{\bcalW^{(l)}}_{\max} \le B,\; \norm{\bb^{(l)}}_{\max} \le B,
\end{aligned}
\end{equation*}
and $\calT_M(\cdot)$ applies truncation operator at level $C$ to each entry of a $d_{L+1}$ dimensional vector, that is, 
$[\calT_M(\calZ)]_{i_1\cdots i_{M_L}} = \sgn(z_{i_1\cdots i_{M_L}})( |z_{i_1\cdots i_{M_L}}| \wedge M)$. 
We denote it as $\calC(L, D_{in}, D_{out}, W, M, B)$ if the width parameter $\bd = (D_{in}, W, W, \cdots, W, D_{out})$, which we referred as deep ReLU network with depth $L$ and width $W$ for brevity. \end{definition}

We now define quantities we aim to bound theoretically. The empirical $\ell_2$ loss is defined as
\begin{equation}
    R_n(f) = \frac{1}{n} \sum_{i=1}^n\paran{y_i - f(\boldsymbol{\calX}_i)}^2
    \equiv \frac{1}{n} \sum_{i=1}^n\paran{y_i - \tilde{f}(\boldsymbol{\calC}_i)}^2,
\end{equation}
where the last equality holds since $\boldsymbol{\calX}$ assumes a Tucker low-rank structure. 
For arbitrary given neural network hyper-parameters $L$ and $W$, we suppose that our TTL estimator is an approximate empirical loss minimizer, that is,
\begin{equation} \label{eqn:erm-obj}
\begin{aligned}
    \hat m(\boldsymbol{\calX}) = \hat f(\boldsymbol{\calC}) 
    \quad\text{with}\quad\\
    R_n(\hat f) \le \inf_{f\in\calC(L, \bar r, 1, W, M, \infty)} R_n(f) 
    + \delta_{opt}
\end{aligned}
\end{equation}
with some optimization error $\delta_{opt}$.
We first present the error bound on the excess risk of the TTL estimator under Tucker low-rankness. 
\begin{theorem} \label{thm:oracle}
Assume tensor feature $\bcalX$ is a $D_1\times\cdots\times D_M$ tensor with low Tucker rank $(R_1, \cdots, R_M)$, $R = \prod_{m=1}^M R_m$. 
Then with probability at least $1-3\exp(-t)$, for large enough $n$, we have
\begin{equation}
\begin{aligned}
    \frac{1}{n}\sum_{i=1}^n\abs{\hat{m}(\boldsymbol{\calX}_i) - m^*(\boldsymbol{\calC}_i)}^2 
    + \EE_{\boldsymbol{\calX},\boldsymbol{\calC}}\brackets{\hat{m}(\boldsymbol{\calX}_i) - m^*(\boldsymbol{\calC}_i)}^2\\
    \le C \paran{\delta_{opt} + \delta_{apr} + \delta_{sto} + \delta_{cor} + t n^{-1}}
\end{aligned}
\end{equation}
for a universal constant $C$ that only depends on $c$ and constants. 
The components in the error bounds are, respectively:\\
NN approximation error
\begin{equation}
    \delta_{apr} = \inf_{f\in\calC(L, \bar r, 1, W, M, \infty)} \norm{f - m^*}_\infty^2
\end{equation}
Stochastic error
\begin{equation}
    \delta_{sto} = (W^2 L^2 + W L R) \log(W L R) \log n / n
\end{equation}
Tensor core error
\begin{equation}
    \delta_{cor} = \sigma^2 \paran{\prod_{m=1}^M D_m}^{-1} \sum_{m=1}^M D_m R_m
\end{equation}

\end{theorem}
Theorem \ref{thm:oracle} establishes a high probability bound on both the out-of-sample mean squared error $\EE_{\boldsymbol{\calX},\boldsymbol{\calC}}\brackets{\hat{m}(\boldsymbol{\calX}_i) - m^*(\boldsymbol{\calC}_i)}$ and in-sample mean squared error $\frac{1}{n}\sum_{i=1}^n\abs{\hat{m}(\boldsymbol{\calX}_i) - m^*(\boldsymbol{\calC}_i)}$. 
The error bound is composed of four terms: 
the optimization error $\delta_{opt}$, 
the neural network approximation error $\delta_{apr}$ to the underlying true regression function $m^*$,
the stochastic error $\delta_{sto}$ scales linearly with $\log n / n$ and $(W^2 L^2 + W L R) \log(W L R)$ which is proportional to the Pseudo-dimension of the neural network class we used, 
and the error $\delta_{cor}$ related to inferring the latent core tensor $\boldsymbol{\calC}$ from the observation $\boldsymbol{\calX}$.
Such an error bound is not applicable without specifying the network hyper-parameters $W$ and $L$. An optimal rate can be further obtained by choosing $W$ and $L$ to trade off the approximation error $\delta_{apr}$ and the stochastic error $\delta_{sto}$. 

The NN approximation error $\delta_{apr}$ can be controlled by the architecture of NN and the optimization error $\delta_{opt}$ can be controlled by optimization algorithm.
Given the depth $L$ and hidden width $W$ of a NN, the stochastic error $\delta_{sto}$ is increasing in $R=\prod_{m=1}^M R_m$, which is the intrinsic dimension of the low-rank weight tensor in the affine transformation \eqref{trl_eq} of a single neuron. 
The ambient dimension of the weight tensor is $D=\prod_{m=1}^M D_m$. 
Under low-rank structure, the intrinsic dimension $R$ is much smaller than the ambient dimension $D$. 
As a result, our proposed {\em Tensor Transformation Layer} (TTL) greatly reduces the stochastic error. 
At the same time, the low-rankness also controls the the core tensor estimation error $\delta_{cor}$ remarkably well. 

The benefit of Tucker low-rankness shows up in the stochastic error $\delta_{sto}$ where $R=\prod_{m=1}^M R_m$ is the total number of elements in the core tensor $\bcalC$ and is also equivalent to the total number of unknown coefficients in {\em low-rank weight} tensor $\boldsymbol{\calW}^{(\ell)}$ in the affine transformation \eqref{trl_eq}. 
The latent Pseudo-dimension of the neural network class we used is reduced thanks to the incorporation of Tucker low-rankness. 

\section{Experiments}

\begin{table*}[ht]
\caption{Performance on molecular and chemical graphs. The best results are given in {\bf bold}. 
\label{classification_result_0_graphs}}
\begin{center}
\resizebox{2.\columnwidth}{!}{
\begin{tabular}{lccccccccccc}
\toprule
\textbf{{Model}} &\textbf{{BZR}} & \textbf{{COX2}} & \textbf{{DHFR}}  & \textbf{D\&D} & \textbf{{MUTAG}} & \textbf{{PROTEINS}} &\textbf{{PTC\_MR}} & \textbf{PTC\_MM} & \textbf{PTC\_FM} & \textbf{PTC\_FR} &\textbf{IMDB\_B}\\
\midrule
CSM~\cite{kriege2012subgraph} & 84.54$\pm$0.65 &79.78$\pm$1.04 & 77.99$\pm$0.96 & OOT &87.29$\pm$1.25 &OOT& 58.24$\pm$2.44 &63.30$\pm$1.70 & 63.80$\pm$1.00 & 65.51$\pm$9.82 & OOT\\
HGK-SP~\cite{morris2016faster} &81.99$\pm$0.30 &78.16$\pm$0.00 & 72.48$\pm$0.65 & 78.26$\pm$0.76 & 80.90$\pm$0.48 &74.53$\pm$0.35& 57.26$\pm$1.41 & 57.52$\pm$9.98 & 52.41$\pm$1.79 & 66.91$\pm$1.46 & 73.34$\pm$0.47\\
HGK-WL~\cite{morris2016faster} &81.42$\pm$0.60 &78.16$\pm$0.00 &  75.35$\pm$0.66 & 79.01$\pm$0.43 & 75.51$\pm$1.34 & 74.53$\pm$0.35& 59.90$\pm$4.30 & 67.22$\pm$5.98 & 64.72$\pm$1.66 &  67.90$\pm$1.81 & 72.75$\pm$1.02\\
WL~\cite{shervashidze2011weisfeiler} &{86.16$\pm$0.97} & 79.67$\pm$1.32 & 81.72$\pm$0.80 & {79.78$\pm$0.36} & 85.75$\pm$1.96 & 73.06$\pm$0.47& 57.97$\pm$0.49 & 67.28$\pm$0.97 & 64.80$\pm$0.85& 67.64$\pm$0.74 & 71.15$\pm$0.47\\
DGCNN~\cite{zhang2018end} &79.40$\pm$1.71 & 79.85$\pm$2.64 & 70.70$\pm$5.00 & {79.37$\pm$0.94}& 85.83$\pm$1.66 & 75.54$\pm$0.94 &  58.59$\pm$2.47 & 62.10$\pm$14.09 & 60.28$\pm$6.67 & 65.43$\pm$11.30 & 70.00$\pm$0.90\\
GCN~\cite{kipf2016semi} &79.34$\pm$2.43& 76.53$\pm$1.82 & 74.56$\pm$1.44  & 79.12$\pm$3.07& 80.42$\pm$2.07 & 70.31$\pm$1.93&62.26$\pm$4.80 & 67.80$\pm$4.00 & 62.39$\pm$0.85 & 69.80$\pm$4.40 & 66.53$\pm$2.33\\
ChebNet~\cite{defferrard2016convolutional} &N/A &N/A &N/A &N/A & 84.40$\pm$1.60 & 75.50$\pm$0.40 &N/A & N/A &N/A & N/A & N/A\\
GIN~\cite{xu2018powerful} &85.60$\pm$2.00 & 80.30$\pm$5.17 & {\bf 82.20$\pm$4.00} & 75.40$\pm$2.60 & 89.39$\pm$5.60 & 76.16$\pm$2.76 & 64.60$\pm$7.00 & 67.18$\pm$7.35 & 64.19$\pm$2.43 & 66.97$\pm$6.17 & 75.10$\pm$5.10\\
g-U-Nets~\cite{gao2019graph} &79.40$\pm$1.20 & 80.30$\pm$4.21 &  69.10$\pm$4.80 & 75.10$\pm$2.20& 67.61$\pm$3.36 & 69.60$\pm$3.50 & 64.70$\pm$6.80 & 67.51$\pm$5.96 & 65.88$\pm$4.26 & 66.28$\pm$3.71 & N/A\\
MinCutPool~\cite{bianchi2020spectral}  & 82.64$\pm$5.05 & 80.07$\pm$3.85 & 72.78$\pm$6.25 & 77.60$\pm$3.10& 79.17$\pm$1.64 & 76.52$\pm$2.58 & 64.16$\pm$3.47 & N/A & N/A  & N/A & 70.77$\pm$4.89 \\
DiffPool~\cite{ying2018hierarchical} &83.93$\pm$4.41 & 79.66$\pm$2.64 &  70.50$\pm$7.80 & 77.90$\pm$2.40& 79.22$\pm$1.02 & 73.63$\pm$3.60 & 64.85$\pm$4.30 & 66.00$\pm$5.36 & 63.00$\pm$3.40 & 69.80$\pm$4.40 & 68.60$\pm$3.10\\
EigenGCN~\cite{ma2019graph}  &83.05$\pm$6.00 & 80.16$\pm$5.80 & N/A  & 75.90$\pm$3.90 & 79.50$\pm$0.66 & 74.10$\pm$3.10 & N/A  & N/A  & N/A & N/A & 70.40$\pm$3.30 \\
CapsGNN~\cite{xinyi2019capsule} &N/A & N/A & N/A & 75.38$\pm$4.17 & 86.67$\pm$6.88 & 76.28$\pm$3.63 & 66.00$\pm$5.90  & N/A  & N/A & N/A & N/A \\
SAGPool~\cite{lee2019self} &82.95$\pm$4.91 & 79.45$\pm$2.98 & 74.67$\pm$4.64 & 76.45$\pm$0.97& 76.78$\pm$2.12 & 71.86$\pm$0.97 &56.41$\pm$1.63  &66.67$\pm$8.57 & 67.65$\pm$3.72 & 65.71$\pm$10.69 & 74.87$\pm$4.09\\
HaarPool~\cite{wang2020haar} &83.95$\pm$5.68 & {82.61$\pm$2.69} & 73.33$\pm$3.72 &  77.40$\pm$3.40 & {90.00$\pm$3.60} dd&73.23$\pm$2.51 & 66.68$\pm$3.22 & 69.69$\pm$5.10 & 65.59$\pm$5.00& 69.40$\pm$5.21 & 73.29$\pm$3.40\\
PersLay~\cite{carriere2020perslay} &82.16$\pm$3.18 &80.90$\pm$1.00 & N/A & N/A &89.80$\pm$0.90 & 74.80$\pm$0.30 & N/A & N/A  & N/A  & N/A & 71.20$\pm$0.70\\
FC-V~\cite{o2021filtration} &85.61$\pm$0.59&81.01$\pm$0.88 &  {81.43$\pm$0.48}  & N/A & 87.31$\pm$0.66&74.54$\pm$0.48& N/A  & N/A & N/A  & N/A & 73.84$\pm$0.36\\
MPR~\cite{bodnar2021deep} &N/A  & N/A & N/A  & N/A & 84.00$\pm$8.60 & 75.20$\pm$2.20 & 66.36$\pm$6.55 & 68.60$\pm$6.30 & 63.94$\pm$5.19& 64.27$\pm$3.78 & 73.80$\pm$4.50\\
SIN~\cite{bodnar2021weisfeiler} &N/A  & N/A & N/A  & N/A & N/A  & {76.50$\pm$3.40} & {66.80$\pm$4.56} & {70.55$\pm$4.79} & {68.68$\pm$6.80} & {69.80$\pm$4.36} & 75.60$\pm$3.20\\
TOGL~\cite{horn2021topological} &N/A & N/A & N/A &  75.70$\pm$2.10 & N/A & 76.00$\pm$3.90 & N/A & N/A & N/A & N/A & N/A\\
\midrule
\textbf{TTG-NN (ours)} &{\bf 87.40 $\pm$ 2.62}
&{\bf 86.73$\pm$3.41} & 78.72$\pm$5.33 & {\bf 80.90$\pm$2.57} & {\bf 93.65$\pm$4.18} &{\bf 77.62$\pm$3.92} &{\bf 68.91$\pm$4.02} & {\bf 74.11$\pm$4.57} & {\bf 69.33$\pm$2.09} & {\bf 73.23$\pm$3.91} & {\bf 76.40$\pm$2.50}\\
\bottomrule
\end{tabular}}
\end{center}
\end{table*}

\subsection{Experiment Settings}
\noindent{\bf Datasets}
We validate our TTG-NN on graph classification tasks using the following real-world chemical compounds, protein molecules, and social network datasets: (i) 4 chemical compound datasets: MUTAG, DHFR, BZR, and COX2, where the graphs are chemical compounds, the nodes are different atoms, and the edges are chemical bonds; (ii) 6 molecular compound datasets: D\&D, PROTEINS, PTC\_MR, PTC\_MM, PTC\_FM, and PTC\_FR, where the nodes represent amino acids and edges represent relationships or interactions between the amino acids, e.g., physical bonds, spatial proximity, or functional interactions; (iii) 1 social network dataset: IMDB-B, where the nodes represent actors/actresses, and edges exist between them if they appear in the same movie. For all graphs, we follow the training principle~\cite{xu2018powerful} and results of the 10-fold cross-validation are reported using standard deviations. OOT indicates out of time (we allow 24 hours for each run).

\noindent{\bf Baselines}
We compare our TTG-NN with 20 state-of-the-art (SOTA) baselines including (1) Comprised of the Subgraph Matching kernel (CSM)~\cite{kriege2012subgraph}, (2) Shortest
Path Hash Graph Kernel (HGK-SP)~\cite{morris2016faster}, (3) Weisfeiler-Lehman Subtree Kernel (HGK-WL)~\cite{morris2016faster}, and (4) Weisfeiler–Lehman (WL)~\cite{shervashidze2011weisfeiler}, (5) Graph Convolutional Network (GCN)~\cite{kipf2016semi}, (6) Chebyshev GCN (ChebNet)~\cite{defferrard2016convolutional}, (7) Graph Isomorphism Network (GIN)~\cite{xu2018powerful}, (8) Deep Graph Convolutional Neural Network (DGCNN)~\cite{zhang2018end}, and (9) Capsule Graph Neural Network (CapsGNN)~\cite{xinyi2019capsule}, (10) GNNs with Differentiable Pooling (DiffPool)~\cite{ying2018hierarchical}, (11) Graph U-Nets (g-U-Nets)~\cite{gao2019graph}, (12) GCNs with Eigen Pooling (EigenGCN)~\cite{ma2019graph}, (13) Self-attention Graph Pooling (SAGPool)~\cite{lee2019self}, (14) Spectral Clustering for Graph Pooling (MinCutPool)~\cite{bianchi2020spectral}, and (15) Haar Graph Pooling (HaarPool)~\cite{wang2020haar}, (16) Topological Graph Neural Networks (TOGL)~\cite{horn2021topological}, (17) PD-based Neural Networks (PersLay)~\cite{carriere2020perslay}, (18) Filtration Curve-based Random Forest (FC-V)~\cite{o2021filtration}, (19) Deep Graph Mapper (MPR)~\cite{bodnar2021deep}, and (20) Simplicial Isomorphism Networ (SIN)~\cite{bodnar2021weisfeiler}.

\noindent{\bf TTG-NN Setup}
We conduct our experiments on one NVIDIA Quadro RTX 8000 GPU card with up to 48GB memory. The TTG-NN is trained end-to-end by using Adam optimizer with the learning rate of \{0.001, 0.01, 0.05, 0.1\}. We use ReLU as the activation function $\sigma(\cdot)$ across our model. The tuning of TTG-NN on each dataset is done via the grid hyperparameter configuration search over a fixed set of choices and the same cross-validation setup
is used to tune the baselines. In our experiments, we set the grid size of $\text{PI}_{\text{Dg}}$ from $20 \times 20$ to $50 \times 50$. The batch size is different for every dataset and ranges from 8 to 128. Each graph convolutional layer and MLP has between 16 and 128 hidden units depending on the dataset regarded. The number of hidden units of TTL are set as \{4, 16, 32\}. We set the layer number of graph convolution blocks as 3, the layer number of MLPs as 2, and choose the dropout ratio as 0.5 for all datasets. We train the models with up to 500 epochs to ensure full convergence and randomly use 50 batches for each epoch. Our code is
available on \href{https://github.com/TaoWen0309/TTG-NN}{GitHub}.

\subsection{Classification Performance}
As shown in Table~\ref{classification_result_0_graphs}, except for DHFR, the performances of our TTG-NN model are significantly better than the runner-ups. More specifically, we found that (i) compared with spectral-based ConvGNNs (i.e., GCN and ChebNet), TTG-NN yields more than
2.20\% relative improvements to the existing best results
for all datasets, (ii) compared with 3 spatial-based ConvGNNs, i.e., DGCNN, GIN, and CapsGNN, TTG-NN achieves a relative gain of up to 16.20\% on all datasets, (iii) TTG-NN outperforms all 6 graph pooling methods (i.e., g-U-Nets, MinCutPool, DiffPool, EigenGCN, SAGPool, and HaarPool) with a significant margin, and (iv) TTG-NN further improves PH-based models and simplicial neural networks by a significant margin on all 11 datasets. To test our TTG-NN's performance on large-scale dataset, we have conducted experiment on the ogbg-molhiv dataset~\cite{hu2020open} with 41,127 graphs. The results in Table~\ref{ogbg-molhiv} show our proposed model is able to achieve promising results on large-scale networks. To sum up,  the results show that our TTG-NN accurately captures the key structural and topological information of the graph, and achieves a highly promising performance in graph classification.

\begin{table}[ht]
    \caption{Graph classification results (\%) on ogbg-molhiv dataset.}
    \label{ogbg-molhiv}
    
    \begin{center}
    \resizebox{1.\columnwidth}{!}{
        \begin{tabular}{lccccc}
        \hline
        Dataset & GCN & GIN & GSN & PNA & TTG-NN (ours) \\
        \hline
        ogbg-molhiv & $75.99 \pm 1.19$ & $77.07 \pm 1.49$ & $77.90 \pm 0.10$ & $79.05 \pm 1.32$ & $\bf{81.50 \pm 0.86}$ \\
        \hline
    \end{tabular}
    }
    \end{center}
\end{table}

\subsection{Ablation Study}

To better understand the importance of different components in TTG-NN, we design the ablation study experiments on MUTAG, BZR, COX2, PTC\_MM, and PTC\_FM. As shown in Table~\ref{ablation_analysis}, if we remove the tensor-view topological convolutional layer (TT-CL), the performance will drop over 8.70\% on average. Specifically, we observe that when removing TT-CL, the performance on graph classification is affected significantly, i.e., TTG-NN outperforms TTG-NN without TT-CL with a relative gain of 12.39\% for PTC\_FM. Moreover, we find that on all 5 datasets, TTG-NN outperforms TTG-NN without the tensor transformation layer (TTL) with an average relative gain of 3.41\%. Furthermore, TTG-NN significantly outperforms TTG-NN without TG-CL on all datasets, which illustrates the importance of learning the tensor-view global structural information. In summary, these results show the effectiveness of both convolutional and topological tensor representation learning for the graph classification problem. In summary, ablating each of the above components leads to performance drops on all datasets compared with the full TTG-NN model, which suggests that the designed components are critical and need to be sufficiently learned.

\begin{table}[ht]
\caption{TTG-NN ablation study.\label{ablation_analysis}}
\begin{center}
\resizebox{1.\columnwidth}{!}{
\begin{tabular}{lccccc}
\toprule
\textbf{{Architecture}} &\textbf{{MUTAG}} & \textbf{{BZR}} & \textbf{{COX2}} & \textbf{{PTC\_MM}} & \textbf{{PTC\_FM}}\\
\midrule
TTG-NN W/o TT-CL& 85.67$\pm$7.80 & 82.73$\pm$3.05 & 78.13$\pm$1.96 &68.67$\pm$7.16 & 60.74$\pm$3.37 \\
TTG-NN W/o TG-CL& 90.60$\pm$3.15 & 86.14$\pm$6.31 & 79.58$\pm$1.84 & 68.20$\pm$7.50 & 62.42$\pm$1.96\\
TTG-NN W/o TTL & 91.22$\pm$5.26 & 85.36$\pm$5.58 & 83.08$\pm$2.49 & 73.80$\pm$5.05 & 64.10$\pm$1.83\\
TTG-NN & {\bf 93.65$\pm$4.18} & {\bf 87.40$\pm$2.62} & {\bf 86.73$\pm$3.41} & {\bf 74.11$\pm$4.57} & {\bf 69.33$\pm$2.09}\\ 
\bottomrule
\end{tabular}}
\end{center}
\end{table}

\subsection{Sensitivity Analysis}

To evaluate the model performance with different tensor decomposition methods, we test the performance of our proposed TTG-NN model with 3 different tensor decompositions, i.e., Tucker, TT, and CP. As Table~\ref{Sensitivity_analysis_decom} shows that (i) on MUTAG, BZR, and COX2, CP method always show better performance than Tucker and TT, and the average relative gain is 5.02\%; (ii) on PTC\_MM and PTC\_FM, we can see that TTG-NN equipped with TT  outperforms TTG-NN with Tucker and CP decompositions respectively. 

\begin{table}[ht]
\caption{Sensitivity analysis of tensor decomposition\label{Sensitivity_analysis_decom}.}
\begin{center}
\resizebox{1.\columnwidth}{!}{
\begin{tabular}{lccccc}
\toprule
\textbf{{Decomposition}} &\textbf{{MUTAG}} & \textbf{{BZR}} & \textbf{{COX2}} & \textbf{{PTC\_MM}} & \textbf{{PTC\_FM}}\\
\midrule
TTL With Tucker & 88.89$\pm$9.94 & 86.91$\pm$4.41 & 79.23$\pm$6.89 & 66.12$\pm$5.52 & 61.03$\pm$3.89\\
TTL With TT & 88.36$\pm$6.37 & 86.91$\pm$3.98 & 82.22$\pm$3.49 & {\bf 74.11$\pm$4.57} & {\bf 69.33$\pm$2.09}\\
TTL With CP & {\bf 93.65$\pm$4.18} & {\bf 87.40$\pm$2.62} & {\bf 86.73$\pm$3.41} & 67.00$\pm$5.23 & 62.45$\pm$5.31\\
\bottomrule
\end{tabular}}
\end{center}
\end{table}


\subsection{Computational Complexity}

The topological complexity of the standard PH matrix reduction algorithm~\cite{edelsbrunner2000topological} runs in time at most $\mathcal{O}(\Xi^3)$, where $\Xi$ is the number of simplices in a filtration. For 0-dimensional PH, it can be computed efficiently using disjoint sets with complexity $\mathcal{O}(\Xi\alpha^{-1}\Xi)$, where $\alpha^{-1}(\cdot)$ is the inverse Ackermann function~\cite{cormen2022introduction}. Furthermore, in Table~\ref{running_time_comparison} we show the running time (i.e., training time per epoch) of the proposed TTL with 3 different tensor decomposition methods on MUTAG, BZR, COX2, PTC\_MM, and PTC\_FM datasets.
{We also compare the running time (training time per epoch; along with the accuracy (\%)) between our TTG-NN model and three runner-ups. Specifically, for MUTAG, TTG-NN: 18.58 seconds (93.65\%) vs. HaarPool: 19.31 seconds (90.00\%) vs. PersLay: 13.00 seconds (89.80\%) vs. GIN 0.38 seconds (89.39\%); for PTC\_MM: TTG-NN 3.97 seconds (74.11\%) vs. SIN: 5.62 seconds (70.55\%) vs. HaarPool: 7.72 seconds (69.69\%) vs. MPR: 9.61 seconds (68.60\%). Compared with runner-ups, TTG-NN always achieves competitive classification performance and computation cost.}

\begin{table}[ht]
\caption{Run time analysis of tensor decomposition(seconds per epoch).\label{running_time_comparison}}
\begin{center}
\resizebox{1.\columnwidth}{!}{
\begin{tabular}{lccccc}
\toprule
\textbf{{Decomposition}} &\textbf{{MUTAG}} & \textbf{{BZR}} & \textbf{{COX2}} & \textbf{{PTC\_MM}} & \textbf{{PTC\_FM}}\\
\midrule
TTG-NN With Tucker & 13.41 s & 44.71 s & 7.52 s & 3.73 s & 4.47 s\\
TTG-NN With TT & 17.58 s & 36.28 s & 7.20 s & 3.97 s & 3.60 s\\
TTG-NN With CP & 18.58 s & 45.02 s & 17.55 s & 12.72 s & 12.85 s\\
\bottomrule
\end{tabular}}
\end{center}
\end{table}

\section{Conclusion}
In this paper, we have proposed a novel {\em Tensor-view Topological Graph Neural Network} (TTG-NN) with graph topological and structural feature tensors. In TTG-NN, TT-CL and TG-CL harness tensor structures to consolidate features from diverse sources, while TTL exploits tensor low-rank decomposition to proficiently manage both model complexity and computational efficiency. TTG-NN architecture can be flexibly extended by incorporating additional graph representation learning modules through the integration of parallel structures with TT-CL and TG-CL. Moreover, we theoretically show that the proposed {\em Tensor Transformation Layer} (TTL) reduces the stochastic noise and error. Extensive experiments on graph classification tasks demonstrate the effectiveness of both TTG-NN and the proposed components. Future research directions include further extending the tensor-view topological deep learning idea to unsupervised/supervised spatiotemporal prediction and community detection.

\noindent{\bf Acknowledgement.} Y.C. has been supported in part by the NASA AIST grant 21-AIST21\_2-0059, and NSF Grant DMS-2335846/2335847.

\clearpage
\bibliographystyle{plain}
\bibliography{main}

\end{document}